%% file: acl2021.tex
\documentclass[11pt,a4paper]{article}
\usepackage[hyperref]{acl2021}
\usepackage{times}
\usepackage{latexsym}

\usepackage{microtype}
\usepackage{enumitem}  % http://ctan.org/pkg/enumitem

\aclfinalcopy % Uncomment this line for the final submission
\input{commands}

\title{Dual Reader-Parser on Hybrid Textual and Tabular Evidence \\ for Open Domain Question Answering}

\author{
 Alexander~Hanbo~Li, 
 Patrick~Ng,
 Peng~Xu,
 Henghui~Zhu, \\
 \textbf{Zhiguo~Wang, Bing~Xiang} \\
 AWS AI Labs, Amazon \\

\texttt{\{hanboli, patricng, pengx, henghui, zhiguow, bxiang\}@amazon.com}\\
}

\date{}

\begin{document}
\maketitle

\begin{abstract}
The current state-of-the-art generative models for open-domain question answering (ODQA) have focused on generating direct answers from unstructured textual information. However, a large amount of world's knowledge is stored in structured databases, and need to be accessed using query languages such as SQL. Furthermore, query languages can answer questions that require complex reasoning, as well as offering full explainability. In this paper, we propose a hybrid framework that takes both textual and tabular evidence as input and generates either direct answers or SQL queries depending on which form could better answer the question. The generated SQL queries can then be executed on the associated databases to obtain the final answers. To the best of our knowledge, this is the first paper that applies Text2SQL to ODQA tasks. Empirically, we demonstrate that on several ODQA datasets, the hybrid methods consistently outperforms the baseline models that only take homogeneous input by a large margin. Specifically we achieve state-of-the-art performance on OpenSQuAD dataset using a T5-\textit{base} model. In a detailed analysis, we demonstrate that the being able to generate structural SQL queries can always bring gains, especially for those questions that requires complex reasoning.
% Our proposed hybrid method significantly outperforms the baseline models that use homogeneous input in terms of exact matching scores, and it can generate SQL queries that are mostly executable. We also provide in-depth model analysis on WikiSQL questions that require different kinds of complex reasoning.
\end{abstract}

\section{Introduction}

Open-domain question answering (ODQA) is a task to answer factoid questions without a pre-specified domain. Recently, generative models \citep{roberts2020much,lewis2020retrieval,min2020ambigqa,izacard2020leveraging} have achieved the state-of-the-art performance on many ODQA tasks. These approaches all share the common pipeline where the first stage is retrieving evidence from the free-form text in Wikipedia. However, a large amount of world's knowledge is not stored as plain text but in structured databases, and need to be accessed using query languages such as SQL. Furthermore, query languages can answer questions that require complex reasoning, as well as offering full explainability. In practice, an ideal ODQA model should be able to retrieve evidence from both unstructured textual and structured tabular information sources, as some questions are better answered by tabular evidence from databases. For example, the current state-of-the-art ODQA models struggle on questions that involve aggregation operations such as counting or averaging. 
% \patricng{Does anyone have citation for this?}
%if a question involves aggregation operations like counting or averaging, then the current state-of-the-art generative models have difficulty on directly predicting the final answer. 

One line of research on accessing databases, although not open domain, is translating natural language questions into SQL queries \citep{zhongSeq2SQL2017,xu2017sqlnet,yu2018spider,guo2019towards,wang2018robust,wang2019rat,yu2018typesql,guo2019content,choi2020ryansql}. These methods all rely on knowing the associated table for each question in advance, and hence are not trivially applicable to the open-domain setting, where the relevant evidence might come from millions of tables. 

In this paper, we provide a solution to the aforementioned problem by empowering the current generative ODQA models with the Text2SQL ability. More specifically, we propose a \textbf{dual reader-parser (\textsc{DuRePa})} framework that can take both textual and tabular data as input, and generate either direct answers or SQL queries based on the context\footnote{Our code is available at \url{https://github.com/awslabs/durepa-hybrid-qa}}. If the model chooses to generate a SQL query, we can then execute the query on the corresponding database to get the final answer. Overall, our framework consists of three stages: retrieval, joint ranking and dual reading-parsing.
% \peng{Let's be consistent here: either we use, retrieval, joint reranking and dual reading-parsing, or retrieval, hybrid reranking and hybrid reading-parsing. Just align with the later subsection titles}. 
% and all of these three components work on hybrid resources\peng{technically, retrieval stage is not hybrid. it happens on two separate indices. I suggest to remove that last sentence.} 
% In the first step, we retrieve supporting candidates of both textual and tabular types. Then a reranker predicts how relevant each candidate is to the question. Finally we use fusion-in-decoder \citep{izacard2020leveraging} for our reader-parser which takes an input all the reranked passages in addition to the question and generates direct answers and SQL queries.
First we retrieve supporting candidates of both textual and tabular types, followed by a \textit{joint} reranker that predicts how relevant each supporting candidate is to the question, and finally we use a fusion-in-decoder model \citep{izacard2020leveraging} for our reader-parser,  which takes all the reranked candidates in addition to the question to generate direct answers or SQL queries. 

To evaluate the effectiveness of our \textsc{DuRePa}, we construct a hybrid dataset that combines SQuAD \citep{rajpurkar2016squad} and WikiSQL \citep{zhongSeq2SQL2017} questions. We also conduct experiments on NaturalQuestions (NQ) \citep{kwiatkowski2019natural} and OTT-QA \citep{chen2020open} to evaluate DuRePa performance. As textual and tabular open-domain knowledge, we used textual and tabular data from Wikipedia via Wikidumps (from Dec. 21, 2016) and Wikitables \citep{bhagavatula2015tabel}. We study the model performance on different kinds of questions, where some of them only need one supporting evidence type while others need both textual and tabular evidence. On all question types, \textsc{DuRePa} performs significantly better than baseline models that were trained on a single evidence type. We also demonstrate that \textsc{DuRePa} can generate human-interpretable SQLs that answer questions requiring complex reasoning, such as calculations and superlatives. 

Our highlighted contributions are as follows:
\begin{itemize}[noitemsep,topsep=0pt,leftmargin=*]
% \item We introduce a task and a dataset of open domain question answering on both textual and tabular evidence data.
% \item We propose a multi-modal generative framework that incorporates hybrid textual and tabular corpus with the Text2SQL ability for ODQA tasks. To the best of our knowledge, this is the first work that investigates Text2SQL in the open-domain QA setting. 
\item We propose a multi-modal framework that incorporates hybrid knowledge sources with the Text2SQL ability for ODQA tasks. To the best of our knowledge, this is the first work that investigates Text2SQL in the ODQA setting.
\item We propose a simple but effective generative approach that takes both textual and tabular evidence and generates either direct answers or SQL queries, automatically determined by the context. With that, we achieve the state-of-the-art performance on OpenSQuAD using a T5-\textit{base} model.
\item We conduct comprehensive experiments to demonstrate the benefits of Text2SQL for ODQA tasks. We show that interpretable SQL generation can effectively answer questions that require complex reasoning in the ODQA setting.
\end{itemize}

\section{Related Work}
\paragraph{Open Domain Question Answering}
ODQA has been extensively studied recently including extractive models \citep{chen2017reading,clark2017simple,wang2019multi,min2019discrete,yang2019end} that predict spans from evidence passages, and generative models \citep{raffel2019exploring,roberts2020much,min2020ambigqa,lewis2020retrieval,izacard2020leveraging} that directly generate the answers. \citet{wang2017r,wang2017evidence,nogueira2019passage} proposed to rerank the retrieved passages to get higher top-n recall.

\paragraph{Table Parsing}
Text2SQL is a task to translate natural questions to executable SQL queries. \citet{brad2017dataset} proposed SENLIDB dataset which only contains 29 tables and lacks annotation in their training set. Recently, with datasets like WikiSQL \citep{zhongSeq2SQL2017}, Spider \citep{yu2018spider} and CoSQL \citep{yu2019cosql} being introduced, many works have shown promising progress on these dataset \citep{yu2018syntaxsqlnet,he2019x,hwang2019comprehensive,min2019discrete,wang2019rat,choi2020ryansql,guo2019towards,lyu2020hybrid,zhang2019editing,zhong2020grounded,shi2020learning}. Another line of work proposes to reason over tables without generating logical forms \citep{neelakantan2015neural,lu2016neural,herzig2020tapas,yin2020tabert}. However, they are all closed-domain and each question is given the associated table.

\paragraph{Hybrid QA}

\citet{chen2020open} also proposed an open-domain QA problem with textual and tabular evidence. Unlike our problem, they generate an answer directly from the tabular evidence instead of generating an SQL query. In addition, they assume some contextual information about table is available during retrieval stage (e.g. their fusion-retriever is pretrained using hyperlinks between tables and paragraphs), whereas we don't use any link information between tables and passages. Moreover, \citet{chen2020hybridqa} proposed a closed-domain hybrid QA dataset where each table is linked to on average 44 passages. Different from ours, their purpose is to study multi-hop reasoning over both forms of information, and each question is still given the associated table.
% Preceding multi-hop works \citep{sun2018open,sun2019pullnet} create hybrid datasets by randomly replacing KB triples with text corpus and find relevant sub-graph that answers a question.

\section{Method}
In this section, we describe our method for hybrid open-domain question answering. It mainly consists of three components: (1) a retrieval system; (2) a joint reranker and (3) a dual Seq2Seq model that uses fusion-in-decoder \citep{izacard2020leveraging} to generate direct answer or SQL query.
\begin{figure*}[htb]
 %\hspace{-0.3in}
 \centering
    \includegraphics[width=0.80\linewidth]{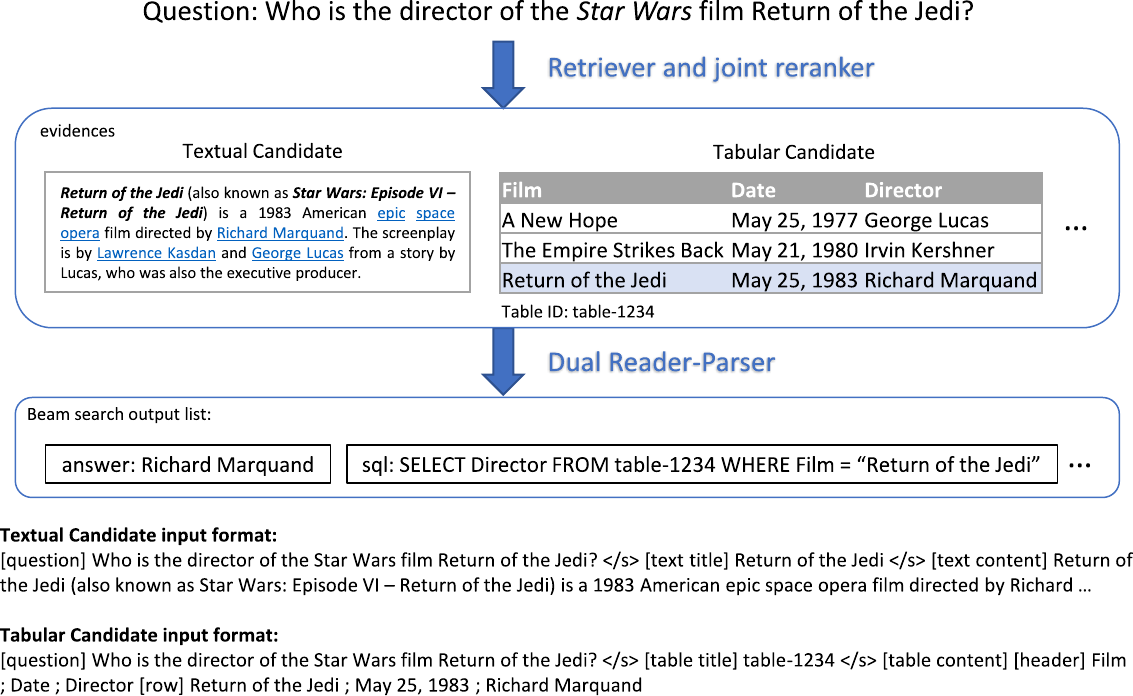}
%\vspace{-0in}
\caption{\label{fig:overall_pipeline}The pipeline of our proposed hybrid model. The candidates are retrieved from knowledge source such as Wikipedia including both paragraphs and tables. 
Then a generative Seq2Seq model reads the question and all the candidates, and produces $k$ outputs using beam search. Each output can be either a final answer or an intermediate SQL query. The types and order of the outputs are automatically determined by the model itself.
% \peng{Should we also include reranker into the pipeline and move the fig on top of second page, probably in a horizontal way?}
}
\end{figure*}

\subsection{Retrieval} 
For the hybrid open-domain setting, we build two separate search indices -- one for textual input and another for tabular input. 
% implemented in Amazon Elastic Search engine\footnote{\url{https://aws.amazon.com/elasticsearch-service/}}.
% \peng{We need to elaborate a bit more about how we build the index of tabular corpus.} 
For paragraphs, we split them into passages of at most 100 words. For tables, we flattened each table into passages by concatenating cell values along each row. If the flattened table exceeds 100 words, we split it into a separate passage, respecting row boundaries. The column headers are concatenated to each tabular passage. Some examples of flattened tables are given in the Appendix \ref{ssec:appendix_bm25_format}.

Given a natural language question, the retrieval system retrieves 100 textual and 100 tabular passages as the support candidates from the textual and tabular indices, respectively, using BM25 \citep{robertson1995okapi} ranking function.

\subsection{Joint Reranking}
The purpose of our reranking model is to produce a score $s_i$ of how relevant a candidate (either an unstructured passage or table) is to a question.
% , and is based on a large-scale pretrained model, BERT \citep{devlin2018bert}. 
%Given a natural language question and a retrieved candidate of either unstructured text or flattened table, %the reranker assigns a relevance score. 
Specifically, the reranker input is the concatenation of \texttt{question}, a retrieved \texttt{candidate-content}, and its corresponding \texttt{title} if available\footnote{Wikipedia passages have page titles, and tables have table titles.}, separated by special tokens shown in Figure \ref{fig:overall_pipeline}. 
% \texttt{[CLS] question [SEP] title [title] candidate-content [SEP]}.
The candidate content can be either the unstructured text or flattened table. We use $\text{BERT}_{base}$ model in this paper. Following \citet{nogueira2019passage}, we finetune the BERT  \cite{devlin2018bert} model using the following loss:
\begin{align}
    L = - \sum_{i \in \mathcal{I}_{pos}} \log(s_i) - \sum_{i \in \mathcal{I}_{neg}} \log (1-s_i).
\end{align}

The $\mathcal{I}_{pos}$ is sampled from all relevant BM25 candidates, and the set $\mathcal{I}_{neg}$ is sampled from all non-relevant BM25 candidates. Different from \citet{nogueira2019passage}, during training, for each question, we sample 64 candidates including one positive candidate and 63 negative candidates, that is, $|\mathcal{I}_{pos}| = 1$ and $|\mathcal{I}_{neg}| = 63$. If none of the 200 
candidates is relevant, we skip the question. During inference, we use the hybrid reranker to assign a score to each of the 200 candidates, and choose the top 50 candidates as the input to the next module -- the reader-parser model. For the top 50 candidates, we choose them from the joint pool of all candidates, according to the scores assigned by the reranker.

\subsection{Dual Reading-Parsing}

Our dual reader-parser model is based on the fusion-in-decoder (FID) proposed in \citet{izacard2020leveraging}, and is initialized using the pretrained T5 \citep{raffel2019exploring} model. The overall pipeline of the reader-parser is shown in Figure \ref{fig:overall_pipeline}. Each retrieved candidate is represented by its title and content, in the following formats:

\paragraph{Textual Candidate} We represent each textual candidate as the concatenation of the passage title and content, appended by special tokens \texttt{[text title]} and \texttt{[text content]} respectively.

\paragraph{Tabular Candidate} In order to represent a structured table as a passage, we first flatten each table into the following format: each flattened table starts with the complete header names and then followed by rows. Figure \ref{fig:overall_pipeline} presents an example for this conversion.
    
% \begin{quote}
%     \textit{[header] Country ; Film title ; Language ; Director [row] Argentina ; The Island ; Spanish ; Alejandro ...} %[row] Austria ; Tales from the Vienna Woods ; German ; Maximilian Schell
% \end{quote}

Finally, a tabular candidate is the concatenation of the table title and content flattened as a passage, appended by special tokens \texttt{[table title]} and \texttt{[table content]} respectively. We use the table ID as the title so that it can be copied to the generated SQL queries by the model.

\paragraph{Prefix of the Targets}
During training, we also add special tokens \texttt{answer:} or \texttt{sql:} to a targeted sentence depending on whether it is a plain text or a SQL query. For those questions that have both textual answer and SQL query annotations (for example, WikiSQL questions), we create two training examples for each question. During inference, the generated outputs will also contain these two special prefixes, indicating which output type the model has generated.

\paragraph{Dual Reader-Parser} Our generative Seq2Seq model has reader-parser duality. During inference, the model reads the question and all the candidates, and produces $k$ outputs using beam search. Each output can be either a final answer or an intermediate SQL query. Depending on the context, the types and order of the outputs are automatically determined by the model itself. All the generated SQL queries will then be executed to produce the final answers. In this paper, we fix $k=3$ and always generate three outputs for each question.

% \newpage
\section{Experiments}
\input{./experiment}

\section{Discussion and Future Work}
Our experiments consistently show that the proposed framework \textsc{DuRePa} brings significant improvement on answering questions using hybrid types of evidence. Especially on the questions that can be answered by both supporting evidence types, our multi-modal method still shows clear advantage over models using single-type knowledge, implying that our approach could figure out the most relevant evidence to answer a question. We also demonstrate that the dual reader-parser is essential to the good performance of \textsc{DuRePa}; the ability of generating both direct answers and structural SQL queries help \textsc{DuRePa} perform much better than \textsc{FiD+} and other baselines on questions that require complex reasoning like counting or averaging. 

We believe that our methods can be improved in two aspects. 
First, our general framework \cref{fig:overall_pipeline} can be improved by a better retrieval system. For example, instead of using BM25, we can use more powerful neural retrieval models~\citep{karpukhin2020dense}. On the hybrid evidence, one can also use an entity linking module to link the entities between the tables and passages \citep{chen2020open} and utilize the structure information for better multi-hop reasoning. 
Second, as we have demonstrated, having the ability of generating structural SQL queries is a very powerful and necessary feature for answering questions that require complex reasoning. Given the limited Text2SQL data and the difficulty of obtaining such SQL supervision, two interesting future work include (1) getting SQL annotations more efficiently and (2) adapting weakly-supervised approaches like discrete EM \citep{min2019discrete} for model training.

% Arguably it is more difficult to obtain annotated SQL queries than just answer text in terms of creating the training data for QA tasks. 
% How to better train Text2SQL more efficiently with less annotation effort
% From the fine-grained error analysis, many errors are due to the generated cell values not existing in the predicted tables. This can be improved by having a entity linking module to link the named entities in the questions to actually cell values in the tables, or by merging small tables with the same headers into a larger one.

% \clearpage
% \newpage

\bibliographystyle{acl_natbib}
\bibliography{anthology,acl2021}

\clearpage
\appendix
\input{./appendix}

\end{document}

%% file: commands.tex
\usepackage{amsmath,bm,bbm,mathtools}
\usepackage[autostyle]{csquotes}
\usepackage{subcaption,multirow}
\usepackage{amssymb,pifont}
\usepackage{url}            % simple URL
\usepackage[scr=boondox]{mathalfa}
\usepackage{booktabs}       % professional-quality tables
\usepackage{amsfonts}       % blackboard math symbols
\usepackage{nicefrac}       % compact symbols for 1/2, etc.
\usepackage{microtype}      % microtypography
\usepackage[colorinlistoftodos]{todonotes}
\usepackage{soul}
\usepackage{xcolor,colortbl}
\usepackage{tabularx}
\usepackage[capitalise]{cleveref}

\definecolor{anti-flashwhite}{rgb}{0.95, 0.95, 0.96}
\definecolor{antiquewhite}{rgb}{0.98, 0.92, 0.84}

\newcolumntype{g}{>{\columncolor{gray}}c}
\newcolumntype{w}{>{\columncolor{white}}c}
\newcolumntype{d}{>{\columncolor{anti-flashwhite}}c}
\newcolumntype{a}{>{\columncolor{antiquewhite}}c}

\def\eqref#1{equation~\ref{#1}}
\def\1{\bm{1}}

\DeclareMathAlphabet{\mathsfit}{\encodingdefault}{\sfdefault}{m}{sl}
\SetMathAlphabet{\mathsfit}{bold}{\encodingdefault}{\sfdefault}{bx}{n}

%% file: experiment.tex
In this section, we report the performance of the proposed method on several hybrid open-domain QA datasets.

\subsection{Datasets}\label{ssec:datasets}
In this section, we describe all the datasets we use in our experiments. First we summarize the statistics of the open-domain QA datasets we use in \cref{tab:data}.
\begin{table}[h]
\centering
\small
\begin{tabular}{lcc}
\toprule
\textbf{Dataset} & \textbf{\#Train\&Dev}& \textbf{\#Test}          \\
\midrule
OpenSQuAD    & 82,599                 & 5,000            \\
OpenNQ       & 87,925                 & 3,610             \\
OTT-QA       & 41,469                 & 2,214            \\
OpenWikiSQL  & 52,026                 & 7,764            \\
Mix-SQuWiki  & 134,625                & 12,764           \\
WikiSQL-both &  --                      & 3,029            \\
\bottomrule
\end{tabular}
\caption{Statistics of Datasets}
\label{tab:data}
\end{table}

\begin{table*}[htb]
\centering
\small
\begin{tabular}{llcccc}
\toprule
\textbf{Model}                                  & \textbf{Evidence Corpus Type }            & \textbf{OpenSQuAD}     & \textbf{OpenNQ}        & \textbf{OTT-QA}        & \textbf{OpenWikiSQL}   \\
\midrule
FiD(T5-\textit{base})             & Text-only              & 53.4          & 48.2          & -             & -             \\
FiD(T5-\textit{large})            & Text-only              & \textbf{56.7} & 51.4          & -             & -             \\
IR+CR                    & Text+Table w/o SQL    & -             & -             & 14.4          & -             \\
FR+CR                    & Text+Table w/o SQL    & -             & -             & \textbf{28.1}\footnotemark & -             \\
Unified Model            & Text+NQ Table w/o SQL & -             & \textbf{54.6}\footnotemark         & -             & -             \\
\midrule
\textit{Ours}            &                        &               &               &               &               \\
\textsc{FiD+}                 & Text-only              & 56.4          & 45.2          & 14.5          & 13.9          \\
\textsc{FiD+}                 & Table-only w/o SQL    & 2.5           & 14.3          & 4.1           & 30.3          \\
\textsc{DuRePa}                    & Table-only with SQL      & 2.7           & 14.8          & 4.7           & 40.2          \\
\textsc{FiD+}                 & Text+Table w/o SQL    & 56.4          & 46.7          & 15.0          & 30.9          \\
\textsc{DuRePa}                    & Text+Table with SQL      & \textbf{57.0} & \textbf{48.0} & \textbf{15.8}     & \textbf{42.6} \\
\bottomrule
\end{tabular}
\caption{Comparison to the state-of-the-art on open-domain QA datasets. The numbers reported are in EM metric. FiD(T5-\textit{base} \& T5-\textit{large}) is reported from \cite{izacard2020leveraging}, IR+CR (Iterative Retrieval+Cross-block Reader) and FR+CR (Fusion Retrieval+Cross-block Reader) are from \cite{chen2020open}, Unified Model is from \cite{oguz2020unified}.
Comparing \textsc{DuRePa}  with \textsc{FiD+}  , we observe that having the ability to generate structural queries is always beneficial even for questions with mostly extractive answers like SQuAD and NQ.}
\label{tab:main}
\end{table*}

\paragraph{OpenSQuAD} is an open-domain QA dataset constructed from the original SQuAD-v1.1 \citep{rajpurkar2016squad}, which was designed for the reading comprehension task, consisting of 100,000+ questions posed by annotators on a set of Wikipedia articles, where the answer to each question is a span from the corresponding paragraph.
\paragraph{OpenNQ} is an open-domain QA datasets constructed from the NaturalQuestions \citep{kwiatkowski2019natural}, which was desgined for the end-to-end question answering task. The questions were from real google search queries and the answers were from Wikipedia articles annotated by humans.  
\paragraph{OTT-QA}\citep{chen2020open} is a large-scale open table-and-text question answering dataset for evaluating open QA over both tabular and textual data. The questions were constructed through ``decontextualization" from HybridQA \cite{chen2020hybridqa} with additional 2200 new questions mainly used in dev/test set. OTT-QA also provides its own corpus which contains over 5 million passages and around 400k tables.
\paragraph{OpenWikiSQL} is an open-domain Text2SQL QA dataset constructed from the original WikiSQL \citep{zhongSeq2SQL2017}. WikiSQL is a dataset of 80,654 annotated questions and SQL queries distributed across 24,241 tables from Wikipedia. 
% In this paper, we truncate and serialize each table into passages not exceeding 100 words, respecting row boundaries.

\paragraph{Mix-SQuWiki} is the union of OpenSQuAD and OpenWikiSQL datasets.

\paragraph{WikiSQL-both} is a subset of OpenWikiSQL evaluation data that contains the questions that can be answered by both textual and tabular evidences. The purpose of this dataset is to study when both types of evidence are possible to answer a question, whether the hybrid model can still choose the better one. We select these questions in a weakly-supervised way by only keeping a question if the groundtruth answer is contained in both textual and tabular BM25 candidates. For example in Figure \ref{fig:overall_pipeline}, the answer ``Richard Marquand" can be found in both types of passages. We filter out some trivial cases where the answer shows up in more than half of the candidates. \footnote{For example, some numerical number like "1" is a very common substring and shows up in most of the candidates.}
% \footnote{\peng{How to filter out trivia cases?}}

\paragraph{Wikipedia Passages and Tables}
For the textual evidences, we process the Wikipedia 2016 dump and split the articles into overlapping passages of 100 words following \citep{wang2019multi}. To create the tabular evidences, we combine 1.6M Wikipedia tables \citep{bhagavatula2015tabel} and all the 24,241 WikiSQL tables, and flatten and split each table into passages not exceeding 100 words, in the same format mentioned in the previous section. We use these two collections as the evidence sources for all the QA datasets except for OTT-QA, where we use its own textual and tabular collections.

\subsection{Implementation Details}

\paragraph{Retriever and Reranker.} We conduct BM25 retrieval using Elasticsearch 7.7 \footnote{https://www.elastic.co/} with the default settings. And we use a BERT reranker initialized with pretrained BERT-\textit{base-uncased} model. 

% \peng{Please add training details here.} 

\paragraph{Dual Reader and Parser with fusion-in-decoder.} Similar to \cite{izacard2020leveraging}, we initialize the fusion-in-decoders with the pretrained T5 model \citep{raffel2019exploring}. We only explore T5-\textit{base} model in this paper, which has 220M parameters. 

For both reranker and FiD models, we use Adam optimizer \citep{kingma2014adam} with a maximum learning rate of $10^{-4}$ and a dropout rate of 10\%. The learning rate linearly warms up to $10^{-4}$ and then linearly anneals to zero. We train models for 10k gradient steps with a batch size of 32, and save a checkpoint every 1k steps. For the FiD model, when there are multiple answers for one question, we randomly sample one answer from the list. 
% Unless otherwise specified, in the single-type input setting, the candidates are either 50 textual passages or 50 tables, while in the hybrid setting, the input consists of 50 hybrid candidates. During training, the candidates are from BM25 and for each question, we randomly sample one positive candidate and $k$ ($k=24$ for single-type input and $k=49$ for hybrid input) negative candidates from the BM25 results. During inference, the candidates are reranked by the hybrid reranker, to guarantee higher recalls. 
For the FiD model, during inference, we generate 3 answers for each question using beam search with beam size 3.

\begin{table*}[bth]
\centering
\small
\begin{tabular}{l|cc|cc|c}
\toprule
               & BM25                  & \cellcolor[HTML]{E7FAFE}Reranker         & BM25                  & \cellcolor[HTML]{E7FAFE}Reranker         & \multicolumn{1}{c}{\cellcolor[HTML]{E7F0FE}Reranker}      \\
\textbf{Index} & \textbf{textual}     & \cellcolor[HTML]{E7FAFE}\textbf{textual} & \textbf{tabular}     & \cellcolor[HTML]{E7FAFE}\textbf{tabular} & \multicolumn{1}{c}{\cellcolor[HTML]{E7F0FE}\textbf{hybrid}} \\
\hline
R@1          & 34.40                & \cellcolor[HTML]{E7FAFE}69.76            & 1.60                 & \cellcolor[HTML]{E7FAFE}10.16            & \cellcolor[HTML]{E7F0FE}\textbf{69.92}                             \\
% Top-2          & 42.40                & \cellcolor[HTML]{E7FAFE}74.40            & 2.52                 & \cellcolor[HTML]{E7FAFE}12.82            & \cellcolor[HTML]{E7F0FE}74.62                             \\
% Top-3          & 46.90                & \cellcolor[HTML]{E7FAFE}76.36            & 3.20                 & \cellcolor[HTML]{E7FAFE}14.84            & \cellcolor[HTML]{E7F0FE}76.74                             \\
% Top-4          & 50.38                & \cellcolor[HTML]{E7FAFE}77.62            & 3.84                 & \cellcolor[HTML]{E7FAFE}15.78            & \cellcolor[HTML]{E7F0FE}78.10                             \\
% Top-5          & 52.58                & \cellcolor[HTML]{E7FAFE}78.42            & 4.22                 & \cellcolor[HTML]{E7FAFE}16.64            & \cellcolor[HTML]{E7F0FE}79.06                             \\
R@10         & 59.38                & \cellcolor[HTML]{E7FAFE}80.30            & 6.34                 & \cellcolor[HTML]{E7FAFE}18.88            & \cellcolor[HTML]{E7F0FE}\textbf{80.90}                             \\
R@25         & 65.92                & \cellcolor[HTML]{E7FAFE}81.64            & 8.84                 & \cellcolor[HTML]{E7FAFE}21.20            & \cellcolor[HTML]{E7F0FE}\textbf{82.42}                             \\
R@50         & 72.16                & \cellcolor[HTML]{E7FAFE}82.50            & 12.36                & \cellcolor[HTML]{E7FAFE}22.62            & \cellcolor[HTML]{E7F0FE}\textbf{83.26}                             \\
R@100        & 76.50                & \cellcolor[HTML]{E7FAFE}83.44            & 15.04                & \cellcolor[HTML]{E7FAFE}23.72            & \cellcolor[HTML]{E7F0FE}\textbf{84.10}                             \\
% Top-200        & 80.48                & \cellcolor[HTML]{E7FAFE}84.02            & 17.80                & \cellcolor[HTML]{E7FAFE}25.76            & \cellcolor[HTML]{E7F0FE}85.18                             \\
% \hline
% MAP            & 28.84                & \cellcolor[HTML]{E7FAFE}49.94            & 2.22                 & \cellcolor[HTML]{E7FAFE}7.52             & \cellcolor[HTML]{E7F0FE}46.48                             \\
% MRR            & 42.86                & \cellcolor[HTML]{E7FAFE}73.59            & 3.15                 & \cellcolor[HTML]{E7FAFE}13.09            & \cellcolor[HTML]{E7F0FE}73.90                             \\
\bottomrule

\end{tabular}
\caption{\label{tab:recall_squad}Recalls on top-$k$ textual, tabular or the hybrid candidates for SQuAD questions. The recalls on hybrid inputs are almost the same as or even better than the best recalls on individual textual or tabular inputs, meaning that the reranker is able to jointly rank both types of candidates and provide better evidences to the next component – the reader-parser.}
\end{table*}

\subsection{Main Results}\label{sec:main_results}

\footnotetext[3]{\citet{chen2020open} uses a fusion-retriever to retrieved table-passages blocks as evidences. To construct the fusion blocks, they train a GPT-2 model using extra hyperlink information to link table cell to passages. In contrast, we do not use any hyperlink information.}
\footnotetext[4]{\citet{oguz2020unified} uses tables provided by NQ training data (less than 500k in total), whereas we use all the tables extracted from Wikipedia dumps (around 1.6M in total).}

We present the end-to-end results on the open-domain QA task comparing with the baseline methods as show in \cref{tab:main}. 

We build models with 5 different settings based on the source evidence modality as well as the format of model prediction. Specifically, we consider single modality settings with only textual evidence or tabular evidence and the hybrid setting with both textual and tabular evidence available. For tabular evidence, the models either predict direct answer text or generate structure SQL queries. Note we also consider a baseline model, \textsc{FiD+}  , a FiD model that only generates direct answer text, but can make use of both textual and tabular evidence.

First, in the single modality setting, we observe that for OpenSQuAD, OpenNQ and OTT-QA datasets, textual QA model is performing significantly better than tabular QA models, while for OpenWikiSQL, it is the opposite. This is expected due to the nature of the construction process of those datasets. In the hybrid setting, the hybrid models outperform single modality models consistently across all these datasets. This indicates hybrid models are more robust and flexible when dealing with questions of various types in practice.

Comparing \textsc{DuRePa}  with \textsc{FiD+}  , we observe that having the ability to generate structural queries is always beneficial even for extractive questions like SQuAD and NQ.  And for WikiSQL-type questions, the gain of SQL generation is significant.

% \peng{Discussing our results with SOTA results}

On OpenSQuAD dataset, our \textsc{DuRePa}  model using hybrid evidences achieves a new state-of-the-art EM score of 57.0. It is worth noting that the previous best score was attained by FiD using T5-\textit{large} model, while our model is using T5-\textit{base}, which has much fewer parameters. On NQ dataset, \textsc{FiD+} with text-only evidences has lower EM score compared with FiD-base, despite having the same underlying model and inputs. We suspect that this is because (1) we truncate all passages into at most 150 word pieces while in FiD paper they keep 250 word pieces, so the actual input (top-100 passages) to our FiD model is much less than that in the FiD paper; and (2) we use BM25 to retrieve the initial pool of candidates instead of trained embedding-based neural retrieval model\citep{karpukhin2020dense,izacard2020leveraging}. 
%we are using Wikipedia dumps 2016 while they use Wikipedia dumps 2018.
Nevertheless, the \textsc{DuRePa}  model with hybrid evidences still improve the EM by 2.8 points compared to \textsc{FiD+} using only text inputs. On OTT-QA questions, our full model also outperforms the IR+CR baseline by 1.4 points. The FR+CR model is using a different setting where they use hyperlinks between tables and passages to train the fusion-retriever (FR), so the result is not directly comparable to ours. We provide more analysis on OTT-QA in the Appendix. On OpenWikiSQL dataset, enabling SQL generation brings more than 10 points improvement on the EM scores. This is because many questions therein require complex reasoning like COUNT, AVERAGE or SUM on the table evidences. We provide more in-depth analysis in Section \ref{ssec:analysis} including some complex reasoning examples in Table \ref{tab:squwiki}.

\section{Analysis}

% In this section, we provide fine-grained analysis on each component of our proposd method on different types of questions.

\subsection{Retrieval and Reranking Performance}
In this section, we investigate the performance of the BM25 retriever and the BERT reranker using top-k recalls as our evaluation metric.

During both training and inference, for each question, the textual and tabular passages are reranked jointly using a single reranker. On the Mix-SQuWiki dataset, we report the reranking results on SQuAD questions in Table \ref{tab:recall_squad}. The result on WikiSQL questions is in Table \ref{tab:recall_wikisql} in Appendix. To provide better insights on the reranker's performance, we show the top-$k$ recalls on textual, tabular and hybrid evidences separately.
% \peng{How many candidates are reranked in each of three cases during inference?}

From Table \ref{tab:recall_squad}, on both textual and tabular candidates, recall@25 of the ranker is even higher than recall@100 of the BM25 retriever. This suggest that during inference, instead of providing 100 BM25 candidates to the fusion-in-decoder (FiD), only 25 reranked candidates would suffice. 
% Fewer candidates could help save memories and hence speed up the inference\peng{Speeding up inference is questionable because you need to count the time of reranking}. 

In Table \ref{tab:recall_wikisql} and \ref{tab:recall_nq} in Appendix, we observe similar trend with top-25 recalls comparable to top-100 recalls on both WikiSQL and NQ questions. Finally, across all datasets, the recalls on hybrid inputs are almost the same as or even better than the best recalls on individual textual or tabular inputs, meaning that the reranker is able to \textit{jointly} rank both types of candidates and provide better evidences to the next component -- the dual reader-parser.

\subsection{Performance of the Reader-Parser}\label{ssec:analysis}
In this section, we discuss the performance of the dual reader-parser on different kinds of questions.

\paragraph{SQL prediction helps with complex reasoning.} In Table \ref{tab:openwikisql}, we compare the top-1 EM execution accuracy of \textsc{DuRePa}  and \textsc{FiD+} on OpenWikiSQL. If \textsc{DuRePa}  generated a SQL, we execute the SQL to obtain its answer prediction. If the ground-truth answer is a list (e.g., What are the names of Simpsons episodes aired in 2008?), we use set-equivalence to evaluate accuracy. \textsc{DuRePa}  outperforms \textsc{FiD+} on the test set in most of the settings. We also compare their performance under a breakdown of different categories based on the ground-truth SQL query. \textsc{DuRePa}  achieved close to 3x and 5x improvements on WikiSQL questions that have superlative (MAX/MIN) and calculation (SUM/AVG) operations, respectively. For COUNT queries, \textsc{FiD+} often predicted either 0 or 1. Thus, these results support our hypothesis that the SQL generation helps in complex reasoning and explainability for tabular question answering. 

\begin{table}[!htb]
\centering
\small
\begin{tabular}{lccc}
\toprule
                  & \textsc{DuRePa}   & \textsc{FiD+}  & \#Test  \\ 
\hline
All               & \textbf{47.1}   & 29.3    & 7764                \\
COUNT $\in$ \{0,1\}            & 78.0   & \textbf{82.9}   & 770                 \\
COUNT $\ge 2$         & \textbf{44.4}   & 0.0           & 9                   \\
MIN/MAX           & \textbf{26.6}   & 9.3     & 654                 \\
SUM/AVG           & \textbf{22.6}   & 4.7     & 314                 \\
Comparison ($<$ or $>$) & \textbf{45.8}   & 32.0    & 939                 \\
AND-condition     & \textbf{53.0}   & 31.8    & 2045                \\
Answer is a list  & \textbf{34.3}     & 0.0           & 160     \\
\hline
Direct answers & \textbf{78.7} &	75.6 & 933 \\
\bottomrule
\end{tabular}
\caption{\label{tab:openwikisql} Comparison of \textsc{DuRePa}  and \textsc{FiD+} on OpenWikiSQL dataset. We compare their accuracy under a breakdown of different categories based on the ground-truth SQL query. ``Direct answers" stands for the questions that \textsc{DuRePa} predicts direct answers. \textsc{DuRePa}  significantly outperforms on questions that require complex reasoning such as superlatives and calculations. }
\end{table}

% \subsubsection{Results on Mix-SQuWiki questions}

\begin{table*}[hbt]
\centering
\small
\begin{tabular}{llccccc}
\toprule
\textbf{Model}    & \textbf{Evidence Corpus Type}          & \begin{tabular}[c]{@{}c@{}}\textbf{\% of SQL} \\ \textbf{Answers}\end{tabular} & \begin{tabular}[c]{@{}c@{}}\textbf{Acc of SQL} \\ \textbf{Answers} (\%)\end{tabular} & \begin{tabular}[c]{@{}c@{}}\textbf{\% of Direct} \\ \textbf{Answers}\end{tabular} & \begin{tabular}[c]{@{}c@{}}\textbf{Acc of Direct} \\ \textbf{Answers} (\%)\end{tabular} & \textbf{EM (Overall)} \\
\midrule
\textsc{FiD+} & Text-only           & 0.0                                                          & -                                                                  & 100.0                                                           & 34.0                                                                  & 34.0         \\
\textsc{FiD+} & Table-only w/o SQL & 0.0                                                          & -                                                                  & 100.0                                                           & 19.3                                                                  & 19.3         \\
\textsc{DuRePa}    & Table-only with SQL & 53.9                                                        & 42.5                                                                & 46.1                                                           &  8.4                                                                 & 26.8         \\
\textsc{FiD+} & Text+Table w/o SQL & 0.0                                                          & -                                                                  & 100.0                                                           & 40.0                                                                  & 40.0         \\
\textsc{DuRePa}    & Text+Table with SQL   & 33.5                                                     & 44.1                                                                   & 66.5                                                          & 49.8                                                                      & 47.9         \\
\bottomrule
\end{tabular}
\caption{Detailed results on Mix-SQuWiki dataset under various settings.}
\label{tab:mix-squwiki}
\end{table*}

\paragraph{Using hybrid evidence types leads to better performance.}

Shown in Table \ref{tab:mix-squwiki} is the model performance on the Mix-SQuWiki questions. As the baseline models, if we only use a single evidence type, the best top-1 EM is 34.0, achieved by the model \textsc{FiD+} using only textual candidates. However, if we use both evidence types, the hybrid model \textsc{DuRePa} attains a significantly better top-1 EM of 47.9, which implies that including both textual and tabular evidences leads a better model performance on Mix-SQuWiki.
Furthermore, we observe that the model \textsc{DuRePa} has a better top-1 EM compared to \textsc{FiD+}, suggesting that the answers for some of these questions need to be obtained by executing SQL queries instead of generated directly. In Table \ref{tab:squwiki}, we samples some questions on which the model \textsc{DuRePa} predicts the correct answers but the model \textsc{FiD+} fails.

\begin{table*}[hbt]
\centering
\small
\begin{tabular}{llccccc}
\toprule
\textbf{Model}    & \textbf{Evidence Corpus Type}          & \begin{tabular}[c]{@{}c@{}}\textbf{\% of SQL} \\ \textbf{Answers}\end{tabular} & \begin{tabular}[c]{@{}c@{}}\textbf{Acc of SQL} \\ \textbf{Answers} (\%)\end{tabular} & \begin{tabular}[c]{@{}c@{}}\textbf{\% of Direct} \\ \textbf{Answers}\end{tabular} & \begin{tabular}[c]{@{}c@{}}\textbf{Acc of Direct} \\ \textbf{Answers} (\%)\end{tabular} & \textbf{EM (Overall)} \\
\midrule
\textsc{FiD+} & Text-only           & 0.0                                                          & -                                                                  & 100.0                                                           & 38.7                                                                  & 38.7         \\
\textsc{FiD+} & Table-only w/o SQL & 0.0                                                          & -                                                                  & 100.0                                                           &  38.4                                                                     &   38.4           \\
\textsc{DuRePa}    & Table-only with SQL   & 38.6                                                         &  30.4                                                            & 61.4                                                          & 57.2                                                                 & 46.8      \\
\textsc{FiD+} & Text+Table w/o SQL & 0.0                                                          & -                                                                  & 100.0                                                           & 43.2                                                                  &  43.2        \\
\textsc{DuRePa}    & Text+Table with SQL   & 39.8                                                    &   35.5                                                           &  60.2                                                     &    64.0                                                                  &   53.6  \\
\bottomrule
\end{tabular}
\caption{Model Performance on WikiSQL-both dataset. The models are trained on Mix-SQuWiki training data.}
\label{tab:wikisql-both}
\end{table*}

\paragraph{What if the questions can be answered by both textual and tabular evidences?}

Table \ref{tab:wikisql-both} shows the model performance on WikiSQL-both dataset. Recall that all these questions in the dataset can be answered by both type of evidence. First of all, the \textsc{DuRePa} model using tabular evidences behaves better than the \textsc{FiD+} model using textual evidences. This implies on WikiSQL questions, using tabular information leads to better answers. Next, when using only one type of evidence, both \textsc{DuRePa} and \textsc{FiD+} models behave significantly worse than their hybrid counterparts. This indicates that the hybrid model can again figure out which evidence type should be used to provide the correct final answer.

% We study the questions in the WikiSQL-both dataset. All these questions can be answered by both types of evidences. The purpose of this experiment is to study when both textual and tabular evidences are plausible, whether the model can pick one answers a question better.

% From Table \ref{tab:wikisql-both}, first of all, the \textsc{DuRePa} model using tabular evidences behaves better than the \textsc{FiD+} model using textual evidences. It means even when both evidences provide the answer, the difficulties in using the evidences are different. On WikiSQL questions, using tabular information leads to better answers.

% Secondly, when using only one type of evidence, the sw-tex and sw-tab models behave significantly worse than the hybrid model. This indicates that the hybrid model can again figure out which evidence type should be used to provide the correct final answer.

\begin{table*}[!hbt]
\centering
\small
\begin{tabularx}{\textwidth}{lX}
\toprule
% \multicolumn{2}{1}{Examples of the SQuWiki questions that are answered correctly by model \textsc{DuRePa}  but incorrectly bymodel }
% \midrule

\textit{\textbf{Question:}}                & Which party won in the election in voting district Kentucky 5?                                                                                                     \\
\textit{\textbf{Groundtruth:}}             & {[}'democratic'{]}                                                                                                                                                 \\
\textit{\textbf{Top-1 generation by \textsc{DuRePa}:}}  & \texttt{sql: SELECT Party FROM table\_1-1342218-17 WHERE District = "Kentucky 5"}                                                                                           \\
\textit{\textbf{Execution result:}}        & {[}'democratic'{]}                                                                                                                                                 \\
\textit{\textbf{Top-1 generation by \textsc{FiD}+}} & answer: republican                                                                                                                                                 \\
% \hline
% \textit{\textbf{Question:}}                & What was the control for the year with a Conservative Party result of 10 (+5)?                                                                                     \\
% \textit{\textbf{Groundtruth:}}             & {[}'labour hold'{]}                                                                                                                                                \\
% \textit{\textbf{Top-1 generation by sw:}}  & \texttt{sql: SELECT Control FROM table\_2-16041438-1 WHERE Conservative Party = "10 (+5)"}                                                                                  \\
% \textit{\textbf{Execution result:}}        & {[}'labour hold'{]}                                                                                                                                                \\
% \textit{\textbf{Top-1 generation by sw--}} & answer: no overall control                                                                                                                                         \\
%                                           &                                                                                                                                                                    \\
\hline
\textit{\textbf{Question:}}                & Which Condition has an unaffected Partial thromboplastin time, Platelet count, and a Prothrombin time?                                                             \\
\textit{\textbf{Groundtruth:}}             & {[}'aspirin', 'uremia', "glanzmann's thrombasthenia"{]}                                                                                                            \\
\textit{\textbf{Top-1 generation by \textsc{DuRePa}:}}  & \texttt{sql: SELECT Condition FROM table\_1-14006-1 WHERE Partial thromboplastin time = "Unaffected" AND Platelet count = "Unaffected" AND Prothrombin time = "Unaffected"} \\
\textit{\textbf{Execution result:}}        & {[}'uremia', "glanzmann's thrombasthenia", 'aspirin'{]}                                                                                                            \\
\textit{\textbf{Top-1 generation by \textsc{FiD}+}} & answer: vitamin k deficiency or warfarin                                                                                                                           \\
\textit{\textbf{Analysis:}}                & \textcolor{blue}{Answer is a list of medical conditions} \\
% \hline
% \textit{\textbf{Question:}}                & What was the date that the decision was Niittymaki, the attendance larger than 19,207, and the record 28–17–5?                                                     \\
% \textit{\textbf{Groundtruth:}}             & {[}'january 31'{]}                                                                                                                                                 \\
% \textit{\textbf{Top-1 generation by sw:}}  & \texttt{sql: SELECT Date FROM table\_2-11902580-6 WHERE Decision = "niittymaki" AND Attendance \textgreater "19,207" AND Record = "28-17-5"}                                \\
% \textit{\textbf{Execution result:}}        & {[}'january 31'{]}                                                                                                                                                 \\
% \textit{\textbf{Top-1 generation by sw--}} & answer: february 31                                                                                                                                                \\
%                                           &                                                                                                                                                                    \\
\hline
% \textit{\textbf{Question:}}                & What is the total amount of gold medals when there were more than 20 silvers and there were 135 bronze medals?                                                     \\
% \textit{\textbf{Groundtruth:}}             & {[}'1'{]}                                                                                                                                                          \\
% \textit{\textbf{Top-1 generation by \textsc{DuRePa}:}}  & \texttt{sql: SELECT COUNT(Gold) FROM table\_2-15428689-2 WHERE Silver \textgreater 20 AND Bronze \textgreater 135}                                                          \\
% \textit{\textbf{Execution result:}}        & {[}1{]}                                                                                                                                                            \\
% \textit{\textbf{Top-1 generation by \textsc{FiD}+}} & answer: 0                                                                                                                                                          \\

\textit{\textbf{Question:}}                & How many Wins have Goals against smaller than 30, and Goals for larger than 25, and Draws larger than 5?                                             \\
\textit{\textbf{Groundtruth:}}             & {[}'3'{]}                                                                                                                                                          \\
\textit{\textbf{Top-1 generation by \textsc{DuRePa}:}}  & \texttt{sql: SELECT COUNT(Wins) FROM table\_2-18017970-2 WHERE Goals against  \textless ~  30 AND Goals for \textgreater ~ 25 AND Draws \textgreater ~ 5} \\
% \textit{\textbf{Top-1 generation by \textsc{DuRePa}:}}  & \texttt{sql: SELECT COUNT(Wins) FROM table\_2-18017970-2 WHERE Goals against  \textless  30 AND Goals for \textgreater 25 AND Draws \textgreater 5}\\
\textit{\textbf{Execution result:}}        & {[}3{]}                                                                                                                                                            \\
\textit{\textbf{Top-1 generation by \textsc{FiD}+}} & answer: 0  \\
\textit{\textbf{Analysis:}}      & \textcolor{blue}{COUNT operation} \\                                                       
\hline

\textit{\textbf{Question:}}                & What is the highest Rd that Tom Sneva had the pole position in?                                             \\
\textit{\textbf{Groundtruth:}}             & {[}'7'{]}                                                                                                                                                          \\
\textit{\textbf{Top-1 generation by \textsc{DuRePa}:}}  & \texttt{sql: SELECT MAX(Rd) FROM table\_1-10706961-2 WHERE Pole Position = "Tom Sneva"} \\
\textit{\textbf{Execution result:}}        & {[}7{]}                                                                                                                                                            \\
\textit{\textbf{Top-1 generation by \textsc{FiD}+}} & answer: 2.0  \\
\textit{\textbf{Analysis:}}      & \textcolor{blue}{MAX operation} \\                                                       
\hline

\textit{\textbf{Question:}}                & Name the average ERP W and call sign of w237br \\
\textit{\textbf{Groundtruth:}}             & {[}110{]}                                                                                                                                                          \\
\textit{\textbf{Top-1 generation by \textsc{DuRePa}:}}  & \texttt{sql: SELECT AVG(ERP W) FROM table\_2-14208614-1 WHERE Call sign = "w237br"} \\
\textit{\textbf{Execution result:}}        & {[}110{]}                                                                                                                                                            \\
\textit{\textbf{Top-1 generation by \textsc{FiD}+}} & answer: 1.0  \\
\textit{\textbf{Analysis:}}      & \textcolor{blue}{AVG calculation} \\

\bottomrule
\end{tabularx}
\caption{\label{tab:squwiki}Examples of the SQuWiki and OpenWikiSQL questions that are answered correctly by model \textsc{DuRePa} but incorrectly by model \textsc{FiD+}.}
\end{table*}

%% file: appendix.tex
\section{Appendix}

\subsection{Tabular Candidates Format for BM25 Retriever}\label{ssec:appendix_bm25_format}
In order to represent a structured table as passages, we flattened each table into passages by concatenating cell values along each row. If the flattened table exceeds 100 words, we split it into a separate passage, respecting row boundaries. The column headers are concatenated to each tabular passage. For example,
    
\begin{table}[htb]
    \centering
    \small
    \begin{tabular}{c|c|c|c}
    \hline
         \textbf{Country} & \textbf{Film title} & \textbf{Language} & \textbf{Director} \\
         \hline
         Argentina & The Island & Spanish & Alejandro \\
         \hline
         Austria & Tales... & German & Maximilian \\
         \hline
    \end{tabular}
    \label{tab:my_label}
\end{table}
\noindent is flattened to be
\begin{quote}
    \textit{[header] Country ; Film title ; Language ; Director [row] Argentina ; The Island ; Spanish ; Alejandro [row] Austria ; Tales from the Vienna Woods ; German ; Maximilian}
\end{quote}

\subsection{EM of the Unified Models}
In this section, we train two unified models DUPERA and \textsc{FiD}+ on the combined training data from all datasets, and then test them on each individual dataset. Similar to the individual-model setting in Section \ref{sec:main_results}, we observe that having the ability to generate structural queries  is  always  beneficial  even  for  extractive questions like SQuAD and NQ. And for WikiSQL-type questions, the gain of SQL generation is significant.
\begin{table*}[htb]
\centering
\small
\begin{tabular}{llcccc}
\toprule
\textbf{Model}                                  & \textbf{Evidence Corpus Type }            & \textbf{OpenSQuAD}     & \textbf{OpenNQ}        & \textbf{OTT-QA}        & \textbf{OpenWikiSQL}   \\
\midrule
FiD(T5-base)             & Text-only              & 53.4          & 48.2          & -             & -             \\
FiD(T5-large)            & Text-only              & \textbf{56.7} & 51.4          & -             & -             \\
IR+CR                    & Text+Table w/o SQL    & -             & -             & 14.4          & -             \\
FR+CR                    & Text+Table w/o SQL    & -             & -             & \textbf{28.1} & -             \\
Unified Model            & Text+NQ Table w/o SQL & -             & \textbf{54.6}*         & -             & -             \\
\midrule
\textit{Our unified model}            &                        &               &               &               &               \\
\textsc{FiD}+                 & Text-only              & 56.2          & 41.6          & 14.5          & 10.1          \\
\textsc{FiD}+                 & Table-only w/o SQL    & 1.6           & 12.9          & 4.0           & 28.3          \\
DUREPA                   & Table-only with SQL      & 1.7           & 13.2          & 4.5           & 40.3          \\
\textsc{FiD}+                 & Text+Table w/o SQL    & 56.2          & 42.3          & 15.1          & 28.5          \\
DUREPA                   & Text+Table w/ SQL      & \textbf{56.7} & \textbf{45.1} & \textbf{15.8}     & \textbf{41.0} \\
\bottomrule
\end{tabular}
\caption{Comparison of the unified model to the state-of-the-art on open-domain QA datasets. The numbers reported are in EM metric. FiD(T5-base \& T5-large) is reported from \cite{izacard2020leveraging}, IR+C (Iterative Retrieval+Cross-block Reader) and FR+CR (Fusion Retrieval+Cross-block Reader) are from \cite{chen2020open}, Unified Model is from \cite{oguz2020unified}.}
\label{tab:main-appendix}
\end{table*}

\subsection{Recalls on WikiSQL-both and NQ datasets}

We present the recall@k for k = 1, 5, 10, 25, 50 and 100 on the OpenWikiSQL and OpenNQ datasets in Table \ref{tab:recall_nq} and \ref{tab:squwiki_appendix}.

\begin{table*}[!htb]
\centering
\small
\begin{tabular}{l|c
>{\columncolor[HTML]{E7FAFE}}c| c
>{\columncolor[HTML]{E7FAFE}}c| 
>{\columncolor[HTML]{E7F0FE}}c }
\toprule
               & AES                   & Reranked              & AES                   & \multicolumn{1}{c}{\cellcolor[HTML]{E7FAFE}Reranked}              & \multicolumn{1}{c}{\cellcolor[HTML]{E7F0FE}Reranked}              \\
\textbf{Index} & \textbf{textual}      & \textbf{textual}      & \textbf{tabular}      & \multicolumn{1}{c}{\cellcolor[HTML]{E7FAFE}\textbf{tabular}}      & \multicolumn{1}{c}{\cellcolor[HTML]{E7F0FE}\textbf{hybrid}}       \\
\hline
Top-1          & 13.10                 & 18.69                 & 51.70                 & 50.24                                                             & 50.28                                                             \\
% Top-2          & 16.65                 & 21.47                 & 58.62                 & 58.19                                                             & 58.32                                                             \\
% Top-3          & 18.30                 & 23.39                 & 62.38                 & 62.74                                                             & 62.82                                                             \\
% Top-4          & 19.24                 & 24.60                 & 64.66                 & 65.55                                                             & 65.61                                                             \\
Top-5          & 20.08                 & 25.61                 & 66.27                 & 68.15                                                             & 68.15                                                             \\
Top-10         & 22.54                 & 28.84                 & 70.93                 & 74.09                                                             & 74.10                                                             \\
Top-25         & 25.24                 & 32.34                 & 75.53                 & 80.91                                                             & 80.89                                                             \\
Top-50         & 29.66                 & 35.39                 & 80.54                 & 84.78                                                             & 84.63                                                             \\
Top-100        & 33.20                 & 38.14                 & 84.14                 & 87.18                                                             & 87.13                                                             \\
% Top-200        & 36.59                 & 39.98                 & 87.07                 & 88.69                                                             & 88.76                                                             \\
\hline
MAP            & 13.15                 & 18.48                 & 47.63                 & 47.92                                                             & 44.93                                                             \\
MRR            & 16.56                 & 22.03                 & 58.49                 & 58.34                                                             & 58.38                                                             \\
\bottomrule                                                        
\end{tabular}
\caption{\label{tab:recall_wikisql}Recalls on top-$n$ textual, tabular or the hybrid candidates for OpenWikiSQL questions.}
\end{table*}

\begin{table*}[!htb]
\centering
\small
\begin{tabular}{l|cc|cc|c}
\toprule
               & BM25                 & \cellcolor[HTML]{E7FAFE}Reranker         & BM25                 & \cellcolor[HTML]{E7FAFE}Reranker         & \cellcolor[HTML]{E7F0FE}Reranker      \\
\textbf{Index} & \textbf{textual}     & \cellcolor[HTML]{E7FAFE}\textbf{textual} & \textbf{tabular}     & \cellcolor[HTML]{E7FAFE}\textbf{tabular} & \cellcolor[HTML]{E7F0FE}\textbf{both} \\
\hline
Top-1          & 23.68                & \cellcolor[HTML]{E7FAFE}51.63            & 9.31                 & \cellcolor[HTML]{E7FAFE}23.52            & \cellcolor[HTML]{E7F0FE}52.91         \\
Top-5          & 47.12                & \cellcolor[HTML]{E7FAFE}71.00            & 20.42                & \cellcolor[HTML]{E7FAFE}34.82            & \cellcolor[HTML]{E7F0FE}72.33         \\
Top-10         & 56.76                & \cellcolor[HTML]{E7FAFE}75.12            & 26.18                & \cellcolor[HTML]{E7FAFE}38.48            & \cellcolor[HTML]{E7F0FE}76.37         \\
Top-25         & 67.73                & \cellcolor[HTML]{E7FAFE}78.42            & 34.43                & \cellcolor[HTML]{E7FAFE}42.24            & \cellcolor[HTML]{E7F0FE}80.03         \\
Top-50         & 74.76                & \cellcolor[HTML]{E7FAFE}79.64            & 40.66                & \cellcolor[HTML]{E7FAFE}44.54            & \cellcolor[HTML]{E7F0FE}81.33         \\
Top-100        & 80.25                & \cellcolor[HTML]{E7FAFE}80.25            & 45.54                & \cellcolor[HTML]{E7FAFE}45.54            & \cellcolor[HTML]{E7F0FE}82.22         \\
% Top-200        & 84.13                & \cellcolor[HTML]{E7FAFE}80.25            & 45.54                & \cellcolor[HTML]{E7FAFE}45.54            & \cellcolor[HTML]{E7F0FE}82.49         \\
\hline
MAP            & 19.32                & \cellcolor[HTML]{E7FAFE}43.87            & 9.98                 & \cellcolor[HTML]{E7FAFE}22.15            & \cellcolor[HTML]{E7F0FE}42.55         \\
MRR            & 34.54                & \cellcolor[HTML]{E7FAFE}59.95            & 14.79                & \cellcolor[HTML]{E7FAFE}28.81            & \cellcolor[HTML]{E7F0FE}61.31         \\
\bottomrule
\end{tabular}
\caption{\label{tab:recall_nq}Recalls on top-$n$ textual, tabular or the hybrid candidates for OpenNQ questions.}
\end{table*}

\subsection{More Examples that Can Only be Answered by DUREPA}
We provide more examples of the SQuWiki questions that can be answered correctly by DUREPA but incorrectly by \textsc{FiD}+.
\begin{table*}[!htb]
\centering
\small
\begin{tabularx}{\textwidth}{lX}
\toprule
% \multicolumn{2}{1}{Examples of the SQuWiki questions that are answered correctly by model DUREPA but incorrectly bymodel }
% \midrule

% \textit{\textbf{Question:}}                & Which party won in the election in voting district Kentucky 5?                                                                                                     \\
% \textit{\textbf{Groundtruth:}}             & {[}'democratic'{]}                                                                                                                                                 \\
% \textit{\textbf{Top-1 generation by \textsc{DuRePa}:}}  & \texttt{sql: SELECT Party FROM table\_1-1342218-17 WHERE District = "Kentucky 5"}                                                                                           \\
% \textit{\textbf{Execution result:}}        & {[}'democratic'{]}                                                                                                                                                 \\
% \textit{\textbf{Top-1 generation by \textsc{DuRePa}--}} & answer: republican                                                                                                                                                 \\
% \hline
\textit{\textbf{Question:}}                & What was the control for the year with a Conservative Party result of 10 (+5)?                                                                                     \\
\textit{\textbf{Groundtruth:}}             & {[}'labour hold'{]}                                                                                                                                                \\
\textit{\textbf{Top-1 generation by \textsc{DuRePa}:}}  & \texttt{sql: SELECT Control FROM table\_2-16041438-1 WHERE Conservative Party = "10 (+5)"}                                                                                  \\
\textit{\textbf{Execution result:}}        & {[}'labour hold'{]}                                                                                                                                                \\
\textit{\textbf{Top-1 generation by \textsc{FiD}+}} & answer: no overall control                                                                                                                                         \\
                                          &                                                                                                                                                                    \\
% \hline
% \textit{\textbf{Question:}}                & Which Condition has an unaffected Partial thromboplastin time, Platelet count, and a Prothrombin time?                                                             \\
% \textit{\textbf{Groundtruth:}}             & {[}'aspirin', 'uremia', "glanzmann's thrombasthenia"{]}                                                                                                            \\
% \textit{\textbf{Top-1 generation by \textsc{DuRePa}:}}  & \texttt{sql: SELECT Condition FROM table\_1-14006-1 WHERE Partial thromboplastin time = "Unaffected" AND Platelet count = "Unaffected" AND Prothrombin time = "Unaffected"} \\
% \textit{\textbf{Execution result:}}        & {[}'uremia', "glanzmann's thrombasthenia", 'aspirin'{]}                                                                                                            \\
% \textit{\textbf{Top-1 generation by \textsc{DuRePa}--}} & answer: vitamin k deficiency or warfarin                                                                                                                           \\
\hline
\textit{\textbf{Question:}}                & What was the date that the decision was Niittymaki, the attendance larger than 19,207, and the record 28–17–5?                                                     \\
\textit{\textbf{Groundtruth:}}             & {[}'january 31'{]}                                                                                                                                                 \\
\textit{\textbf{Top-1 generation by \textsc{DuRePa}:}}  & \texttt{sql: SELECT Date FROM table\_2-11902580-6 WHERE Decision = "niittymaki" AND Attendance \textgreater "19,207" AND Record = "28-17-5"}                                \\
\textit{\textbf{Execution result:}}        & {[}'january 31'{]}                                                                                                                                                 \\
\textit{\textbf{Top-1 generation by \textsc{FiD}+}} & answer: february 31                                                                                                                                                \\
                                          &                                                                                                                                                                    \\
% \hline
% \textit{\textbf{Question:}}                & What is the total amount of gold medals when there were more than 20 silvers and there were 135 bronze medals?                                                     \\
% \textit{\textbf{Groundtruth:}}             & {[}'1'{]}                                                                                                                                                          \\
% \textit{\textbf{Top-1 generation by \textsc{DuRePa}:}}  & \texttt{sql: SELECT COUNT(Gold) FROM table\_2-15428689-2 WHERE Silver \textgreater 20 AND Bronze \textgreater 135}                                                          \\
% \textit{\textbf{Execution result:}}        & {[}1{]}                                                                                                                                                            \\
% \textit{\textbf{Top-1 generation by \textsc{DuRePa}--}} & answer: 0                                                                                                                                                          \\
\bottomrule
\end{tabularx}
\caption{\label{tab:squwiki_appendix}Examples of the SQuWiki questions that are answered correctly by model \textsc{DuRePa} but incorrectly by model \textsc{\textsc{FiD}+}.}
\end{table*}

\subsection{Zero-shot Performance on OpenNQ}

In Table \ref{tab:nq_examples_appendix}, we present some predictions on NQ questions under the zero-shot setting. The models used here are trained on MixSQuWiki dataset and tested directly on NQ questions without finetuning. 

We observe that for all these questions, the generated SQLs are reasonable and are coherent to what the original questions are asking about. For the first question, even though the groundtruth set is not exactly equal to the executed results, they largely overlap. For the second question, the generated SQL query is indeed searching for colleges that Johnny Manziel has played for. The generated SQL can also be successfully executed and returns a long list of answers, which indeed contains the correct answer "Texas A\&M". In this case, the error comes from an erroneous table. This is also the case for the third example. For the fourth to eighth examples, our model generates SQLs that execute to correct answers. Some of them requires generating multiple conditions connected by "AND". This demonstrates that the DUREPA model indeed learns useful generalizable semantics of SQL queries.

\begin{table*}[!htb]
\centering
\small
\begin{tabularx}{\textwidth}{lX}
\toprule
\textit{\textbf{Question:}}         & cast of law \& order special victim unit                                                                                                                                                                                                                                                                                                                          \\
\textit{\textbf{Groundtruth:}}      & {[}'kelli giddish', 'richard belzer', 'stephanie march', 'diane neal', 'ice-t', 'danny pino', 'dann florek', 'tamara tunie', 'michaela mcmanus', 'mariska hargitay', 'adam beach', 'b. d. wong', 'christopher meloni', 'raúl esparza', 'michelle hurd', 'peter scanavino'{]}                                                                                                                                                                                                                                                                                                                                          \\
\textit{\textbf{Top-1 generation:}} & \texttt{sql: SELECT Cast FROM table\_22266670-7 WHERE Program = "law \& order: special victims unit"}                                                                                                                                                                                                                                                                                      \\
\textit{\textbf{Execution result:}} & {[}'chris orbach', 'donnetta lavinia grays', 'nicholas gonzalez', 'robert john burke', 'kathryn erbe', 'jerry orbach', 'joel de la fuente', 'john schuck', 'william h. burns', 'dean winters', 'connie nielsen', 'adam beach', 'adam baldwin', 'jesse l. martin'{]}                                                                                                                                                                                                                                                                                                                                           \\
\multicolumn{2}{l}{\textit{\textbf{}}}    \\
\hline
\textit{\textbf{Question:}}         & who did johnny manziel play college football for                                                                                                                                                                                                                                                                                                                    \\
\textit{\textbf{Groundtruth:}}      & answer: texas a\&m                                                                                                                                                                                                                                                                                                                                                  \\
\textit{\textbf{Top-1 generation:}} & \texttt{sql: SELECT College(s) played for FROM table\_3401335-11 WHERE Player = "johnny manziel"}                                                                                                                                                                                                                                                                            \\
\textit{\textbf{Execution result:}} & {[}'montana', 'ucla', 'texas a\&m', 'california', 'baylor', 'purdue', 'clemson', 'michigan', 'southern california', 'princeton', 'california lutheran', 'cornell', 'oregon', 'columbia', 'yale , rutgers', 'trinity college', 'tennessee', 'texas', 'iowa', 'kentucky state', 'army', 'ohio state', 'mississippi', 'lsu', 'penn state', 'boston college', 'yale'{]} \\
\multicolumn{2}{l}{\textit{\textbf{}}}                                                                                                                                                                                                                                                                                                                                                                    \\
\hline
\textit{\textbf{Question:}}         & who is the original singer of you re going to love me                                                                                                                                                                                                                                                                                                               \\
\textit{\textbf{Groundtruth:}}      & answer: jennifer holliday                                                                                                                                                                                                                                                                                                                                           \\
\textit{\textbf{Top-1 generation:}} & \texttt{sql: SELECT Original artist FROM table\_30996994-1 WHERE Song (original artist) = "you re going to love me"}                                                                                                                                                                                                                                                         \\
\textit{\textbf{Execution result:}} & {[}'whitney houston', 'diana ross', 'dreamgirls', 'blu cantrell', 'the beatles', 'faith hill', 'jennifer holliday', 'monica'{]}                                                                                                                                                                                                                                     \\
\multicolumn{2}{l}{\textit{\textbf{}}}                                                                                                                                                                                                                                                                                                                                                                    \\
\hline
\textit{\textbf{Question:}}         & who wrote the song what child is this                                                                                                                                                                                                                                                                                                                               \\
\textit{\textbf{Groundtruth:}}      & answer: william chatterton dix                                                                                                                                                                                                                                                                                                                                      \\
\textit{\textbf{Top-1 generation:}} & \texttt{sql: SELECT Author FROM table\_36169771-1 WHERE Song = "what child is this?"}                                                                                                                                                                                                                                                                                        \\
\textit{\textbf{Execution result:}} & {[}'william chatterton dix'{]}                                                                                                                                                                                                                                                                                                                                      \\
\multicolumn{2}{l}{\textit{\textbf{}}}                                                                                                                                                                                                                                                                                                                                                                    \\

\hline\textit{\textbf{Question:}}         & who sings the pokemon theme song season 1                                                                                                                                                                                                                                                                                                                           \\
\textit{\textbf{Groundtruth:}}      & answer: jason paige                                                                                                                                                                                                                                                                                                                                                 \\
\textit{\textbf{Top-1 generation:}} & \texttt{sql: SELECT Vocalist FROM table\_2144389-13 WHERE Title = "pokémon theme" AND Episodes used 1 = "pokémon theme"}                                                                                                                                                                                                                                                     \\
\textit{\textbf{Execution result:}} & {[}'jason paige'{]}                                                                                                                                                                                                                                                                                                                                                 \\
\hline
\textit{\textbf{Question:}}         & when did david akers kick the 63 yard field goal                                                                                                                                                                                                                                                                                                                    \\
\textit{\textbf{Groundtruth:}}      & answer: september 9, 2012                                                                                                                                                                                                                                                                                                                                           \\
\textit{\textbf{Top-1 generation:}} & \texttt{sql: SELECT Date FROM table\_8378967-1 WHERE Distance = "63 yards" AND Kicker = "david akers"}                                                                                                                                                                                                                                                                       \\
\textit{\textbf{Execution result:}} & {[}'september 9, 2012'{]}                                                                                                                                                                                                                                                                                                                                           \\
\hline
\textit{\textbf{Question:}}         & what album is sacrifice by elton john on                                                                                                                                                                                                                                                                                                                            \\
\textit{\textbf{Groundtruth:}}      & answer: sleeping with the past.                                                                                                                                                                                                                                                                                                                                     \\
\textit{\textbf{Top-1 generation:}} & \texttt{sql: SELECT Album FROM table\_4105885-1 WHERE Artist = "elton john" AND Song = "sacrifice"}                                                                                                                                                                                                                                                                          \\
\textit{\textbf{Execution result:}} & {[}'sleeping with the past'{]}                                                                                                                                                                                                                                                                                                                                      \\
\hline
\textit{\textbf{Question:}}         & who played raquel in only fools and horses                                                                                                                                                                                                                                                                                                                          \\
\textit{\textbf{Groundtruth:}}      & answer: tessa peake-jones                                                                                                                                                                                                                                                                                                                                           \\
\textit{\textbf{Top-1 generation:}} & \texttt{sql: SELECT Actor FROM table\_6994109-1 WHERE Role = "raquel" AND Film/Show = "only fools and horses"}                                                                                                                                                                                                                                                               \\
\textit{\textbf{Execution result:}} & {[}'tessa peake-jones'{]}                                                                                         \\
% \hline
% \textit{\textbf{Question:}}         & who played raquel in only fools and horses                                                                                                                                                                                                                                                                                                                          \\
% \textit{\textbf{Groundtruth:}}      & answer: tessa peake-jones                                                                                                                                                                                                                                                                                                                                           \\
% \textit{\textbf{Top-1 generation:}} & \texttt{sql: SELECT Actor FROM table\_6994109-1 WHERE Role = "raquel" AND Film/Show = "only fools and horses"}                                                                                                                                                                                                                                                               \\
% \textit{\textbf{Execution result:}} & {[}'tessa peake-jones'{]}                                                                                         \\
\bottomrule
\end{tabularx}
\caption{\label{tab:nq_examples_appendix}Some predictions on NQ questions under zero-shot learning setting.}
\end{table*}

\subsection{Upper bound analysis on OTT-QA}

In this section, we provide an upper bound analysis for the OTT-QA questions. The main purpose of this experiment is to investigate the main bottleneck of DUREPA model on this multi-hop datasets.

Under the oracle setting, if the groundtruth supporting table is successfully retrieved by the retriever, we then add its hyperlinked supporting passages. Similarly, we add the linked supporting tables for the groundtruth passages if they are successfully retrieved. This is in order to mimic the functionality of the fusion-retriever, which is trained using the hyberlinks in the OTT-QA paper.

The results are presented in Table \ref{tab:ott-qa-upb}. We can see that if we have an "oracle" retriever, the end-to-end EM can be improved to 32.2, which is more than doubled compared to 15.8 EM under the normal setting. Therefore, the retriever is indeed the bottleneck of our method on multi-hop QA. Actually, this also holds true for the baseline models. The FR+CR model also significantly improves upon IR+CR model by only replacing the iterative retriever with the fusion-retriever, which is trained to link each table with its hyperlinked paragraphs.

\begin{table*}[htb]
\centering
\small
\begin{tabular}{llcc}
\toprule
\textbf{Model}    & \textbf{Evidence Corpus Type}          & \textbf{EM (normal setting)} & \textbf{EM (oracle-retriever setting)} \\
IR + CR  & Text+Table w/o SQL    &   14.4  &  -- \\
FR + CR  & Text+Table w/o SQL    &   28.1  &  -- \\
\midrule
\textsc{FiD}+ & Text-only             &   14.5  &  31.9 \\
\textsc{FiD}+ & Table-only w/o SQL    &    4.1  &  3.6  \\            
DUREPA   & Table-only with SQL   &    4.7  &  4.8  \\
\textsc{FiD}+ & Text+Table w/o SQL    &   15.0  &  28.5 \\
DUREPA   & Text+Table with SQL   &   15.8  &  32.2 \\
\bottomrule
\end{tabular}
\caption{Upper bound results on OTT-QA dataset. The results show that the main bottleneck of DUREPA methods on OTT-QA dataset is the retriever.}
\label{tab:ott-qa-upb}
\end{table*}

\subsection{Some Ambiguous WikiSQL Questions}
In Table \ref{tab:ambiguous_errors}, we demonstrate that some errors our models make on WikiSQL questions are actually due to the ambiguity. These questions often have different possible answers depending on the contexts. Nevertheless, the SQL queries generated by DUREPA are reasonable and reflect what the original questions are asking for.

\begin{table*}[!htb]
\centering
\small
\begin{tabularx}{\textwidth}{lX}
\toprule
\textit{\textbf{Question:}}                & What is the record when the opponent is washington redskins?                                                                                                 \\
\textit{\textbf{Groundtruth:}}             & {[}"0–3"{]}                                                                                                                                                 \\
\textit{\textbf{Groundtruth SQL:}}  & \texttt{SELECT Record FROM table\_1-18847692-2 WHERE Opponent = \"Washington Redskins\"}                                                                                           \\
\textit{\textbf{Top-1 generation by \textsc{DuRePa}:}}  & \texttt{SELECT Record FROM table\_2-15581223-3 WHERE Opponent = \"washington redskins\"}                                                                                           \\
\textit{\textbf{Execution result:}}        & {[}"1–0"{]}                                                                                                                                                 \\
\hline

\textit{\textbf{Ambiguity:}}                  & There are many records when the opponent is Washington Redskins. \\
\hline
\textit{\textbf{Question:}}                & With the nickname the swans, what is the home ground?                                                                                                 \\
\textit{\textbf{Groundtruth:}}             & {[}"lilac hill park"{]}                                                                                                                                                 \\
\textit{\textbf{Groundtruth SQL:}}  & \texttt{SELECT Home ground(s) FROM table\_1-18752986-1 WHERE Nickname = \"Swans\"}                                                                                           \\
\textit{\textbf{Top-1 generation by \textsc{DuRePa}:}}  & \texttt{SELECT Home ground(s) FROM table\_2-17982112-1 WHERE Nickname = \"swans\"}                                                                                           \\
\textit{\textbf{Execution result:}}        & {[}""{]}                                                                                                                                                 \\
\hline
\textit{\textbf{Ambiguity:}}                  & There are more than one home grounds with nickname the swans. \\
\hline

\textit{\textbf{Question:}}                & Who had the fastest lap in the Belgian Grand Prix                                                                                                 \\
\textit{\textbf{Groundtruth:}}             & {[}"rubens barrichello"{]}                                                                                                                                                 \\
\textit{\textbf{Groundtruth SQL:}}  & \texttt{SELECT Fastest Lap FROM table\_1-1132600-3 WHERE Grand Prix = \"Belgian Grand Prix\"}                                                                                           \\
\textit{\textbf{Top-1 generation by \textsc{DuRePa}:}}  & \texttt{SELECT Fastest Lap FROM table\_1-1140077-2 WHERE Race = \"Belgian Grand Prix\"}                                                                                           \\
\textit{\textbf{Execution result:}}        & {[}"carlos reutemann"{]}                                                                                                                                                 \\
\hline
\textit{\textbf{Ambiguity:}}                  & The question does not specify which year. Many answers are possible. \\

\bottomrule
\end{tabularx}
\caption{\label{tab:ambiguous_errors}Examples of ambiguous questions in OpenWikiSQL.}
\end{table*}